\documentclass[letterpaper, 10 pt, conference]{ieeeconf}  

\usepackage{fancyhdr}

\IEEEoverridecommandlockouts                              

\usepackage{amsmath} 
\usepackage{graphicx}
\usepackage{paralist}
\usepackage[nameinlink]{cleveref}

\usepackage[utf8]{inputenc}
\usepackage{xcolor}

\usepackage{graphviz}

\newcommand{\ph}[1]{{\textbf{#1:}}}

\usepackage{import}
\graphicspath{{figures/}}

\title{\LARGE \bf Copiloting Autonomous Multi-Robot Missions:\\ A Game-inspired Supervisory Control Interface}

\author{Marcel Kaufmann$^{1,2}$, Robert Trybula$^{3}$, Ryan Stonebraker$^{2}$, Michael Milano$^{2}$, Gustavo J. Correa$^{4}$,\\Tiago S. Vaquero$^{2}$, Kyohei Otsu$^{2}$, Ali-akbar Agha-mohammadi$^{2}$, and Giovanni Beltrame$^{1}$
\thanks{$^{1}$M.K. and G.B. are with
        Polytechnique Montreal, H3T 1J4 Montreal, QC, Canada
        {\tt\small firstname.lastname@polymtl.ca}}%
\thanks{$^{2}$M.K., M.M., R.S., T.S.V., K.O., and A.A. are with NASA Jet Propulsion         Laboratory, California Institute of Technology,
        Pasadena, CA 91109, USA
        {\tt\small firstname.lastname@jpl.nasa.gov}}%
\thanks{$^{3}$R.T. is with University of Southern California, Los Angeles, CA 90007, USA
        {\tt\small trybula@usc.edu}}%
\thanks{$^{4}$G.J.C. is with University of California, Riverside,
CA 92521, USA
        {\tt\small gcorr003@ucr.edu}}%
\thanks{This work has been submitted to the IEEE for possible publication. Copyright may be transferred without notice.}
}

\begin{document}
\bstctlcite{IEEEexample:BSTcontrol}

\maketitle
\thispagestyle{empty}
\pagestyle{empty}

\begin{abstract}

Real-world deployment of new technology and capabilities can be daunting. The recent DARPA Subterranean (SubT) Challenge, for instance, aimed at the advancement of robotic platforms and autonomy capabilities in three one-year development pushes. While multi-agent systems are traditionally deployed in controlled and structured environments that allow for controlled testing (e.g., warehouses), the SubT challenge targeted various types of unknown underground environments that imposed the risk of robot loss in the case of failure. In this work, we introduce a video game-inspired interface, an autonomous mission assistant
and test and deploy these using a heterogeneous multi-agent system in challenging environments.
This work leads to improved human-supervisory control for a multi-agent system reducing overhead from application switching, task planning, execution, and verification while increasing available exploration time with this human-autonomy teaming platform.
\end{abstract}

\section{Introduction}
\ph{Autonomous Exploration and SubT} Robotic exploration and the advancement of autonomy offer new ways to explore potentially dangerous and hard-to-access underground environments. Multi-agent systems have matured in controlled and structured environments like warehouses, factories, and laboratories, while current robotic challenges seek to advance these technologies for search and rescue scenarios, planetary prospecting, and subsurface exploration \cite{asada2019robocup, Hambuchen2017NASA_Rob_Challenge, Link2021_ESA_ESRIC}. Motivated by the search for life on other planets, NASA JPL's team CoSTAR \cite{agha2021nebula} took part in the Defense Advanced Research Projects Agency's (DARPA) Subterranean Challenge (SubT) seeking to advance robotic multi-agent systems and their technology readiness for potential future missions. If brought to other planets (e.g. Mars), subsurface missions could bring new insights into their geologic past as well as on their potential for supporting life in the environmentally protected undergrounds \cite{Titus2021}. In contrast to traditional exploration missions where a team of operators and scientists controls one rover, SubT introduced the challenging requirement that only \textit{a single human supervisor} can  directly interface with the deployed multi-agent team in real-time and when a communication link is established. SubT is divided into three, one year development pushes with major field testing demonstrations. This work focuses on the advancements in our supervisor autonomy and game-inspired user interface that were developed under the restrictions of a worldwide pandemic and deployed during the SubT final competition comprising two preliminary missions (P1 and P2) and the final prize run (F).

\begin{figure}[t]
\centering
  \includegraphics[width=0.45\textwidth]{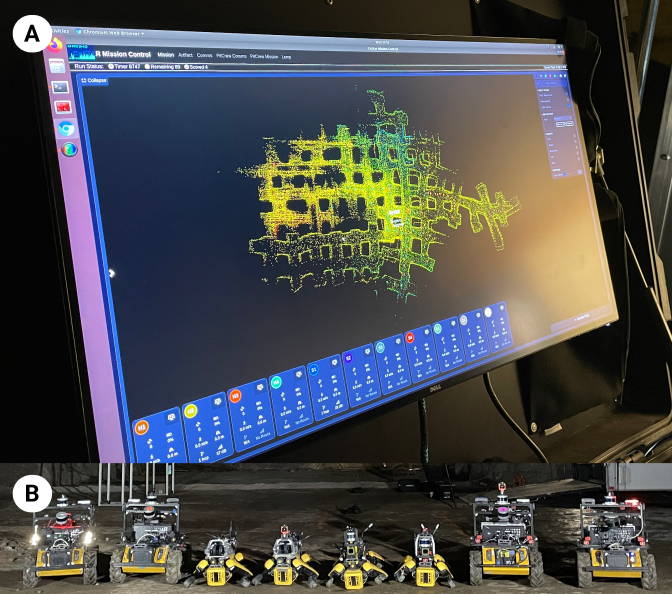}
  \caption{Team CoSTAR's Mission Control user interface (A). (B) a subset of CoSTAR's ground robots showing four customized Boston Dynamic's Spot and Clearpath Husky powered by JPL's autonomy platform NeBULA. Typically a deployment of 4 to 6 ground vehicles was targeted during SubT, but the number of agents is extendable (e.g., see A with 11 robots).}
  \label{fig:robot_team}
\end{figure}

\ph{Human-Robot Collaboration} Achieving man-computer symbiosis \cite{Licklider1960ManComputerSymbiosis} has been a long-time goal of the community to promote a close coupling of human and machine capabilities and ultimately inspire the evolving field of human-robot interaction \cite{chen2021human}. 
This work improves collaborative human multi-robot exploration and search performance fusing our extended autonomy assistant Copilot \cite{Kaufmann2020copilotMike} that uses automated planning techniques with a game-inspired interface design for effective robot deployment, operations, and single operator supervision to create a more symbiotic interaction.  

We present key design choices that are breaking away
from common robot interfacing strategies that were deployed in similar challenge
contexts \cite{kohlbrecher2015human,cerberus} and used interfaces based on the
Robot Operating System's (ROS) visualization tool RViz. Further, we leverage
human-robot interdependencies to inform the design and development of supervised
autonomy and interaction paradigms to achieve our set interaction objectives.
The latest results from the SubT competition ``Finals'' are compared to a
baseline from previous competition runs, namely the ``Urban Circuit'', which
deployed earlier interface and system implementations and interaction paradigms
that we improve with our combined game-inspired interface and enhanced
supervisory autonomy.

In \Cref{sec:related_work}, we look at related work from human-robot interaction
and user interface design. \Cref{sec:problem_requirements_objectives} outlines
the SubT requirements and interaction objectives for our multi-agent exploration
scenario.
We then describe our supervisory autonomy Copilot and its latest implementation
in \Cref{sec:supervised_autonomy}, while \Cref{sec:interface} outlines the user
interface components and human-robot interaction capabilities. Finally, we
present results from our development pushes in \Cref{sec:results} and close with
conclusions and an outlook on future work in
\Cref{sec:conclusions_future_work}.


\section{Related Work}
\label{sec:related_work}
\ph{Human-Robot Interaction and Interface Design} More than sixty years after
the introduction of man-computer symbiosis by
\textit{Licklider}~\cite{Licklider1960ManComputerSymbiosis}, \textit{Chen and
  Barnes}~\cite{chen2021human} conclude that the boundaries of long-term
human-robot symbiosis are still to be pushed by interdisciplinary
collaborations. \textit{Szafir and Szafir}~\cite{Szafir2021HRI_DATA_VIZ} have
identified best practices in the field of data visualization as a key driver to
advance both HRI and data visualization. Complex visualizations and renderings
have become achievable with off-the-shelf hardware, which allows the integration
of visualization principles such as sensemaking~\cite{Szafir2021HRI_DATA_VIZ}
that helps a human digest information. In human-space systems \textit{Rahmani
  et al.}~\cite{rahmani2019space} identified that interface technologies are
currently in development, but their technology readiness levels are not very
mature. Multiple design methods have been introduced in the literature, for
instance, Coactive Design~\cite{coactive_design} which is a structured
approach to analyze human and robot requirements and was used in the context of
the 2015 DARPA Virtual Robotics Challenge that aimed at advancing disaster
response capabilities. \textit{Roundtree et
  al.}~\cite{Roundtree2019VizDesignCollectiveTeams} found that abstract
interface designs that visualize collective status over single agent
information could increase performance; however such designs depend on the task
at hand, team size and mission goals~\cite{chen2021human}. A common testing
strategy in computer game development is
Playtesting~\cite{Wallner2019Playtesting},
which is comparable to simulation and field testing in the multi-robot domain.
The game-inspired development technique RITE, which was introduced in the
context of interface development for the computer game Age of
Empires~\cite{medlock2002usingRITE_AgeOfEmpires}, was used and adapted for fast
development sprints. Additionally, we drew inspiration from real-time strategy games like Age of Empires,
which guided the design of the 3D portion of the interface.

\ph{Robot Challenge Interfaces} During 2013's DARPA Robotics Challenge, team
ViGIR leveraged ROS to control a humanoid robot. The team decided to implement
their interfaces using RViz and built an Operation Control Center consisting of
at least six screens. Robot challenges are found to typically influence
human-robot interaction design and interfaces~\cite{Szafir2021HRI_DATA_VIZ} and
for DARPA's SubT teams, the common design practice was based on RViz and ROS
plugins (\cite{csiro,cornellSubTJFR,scherer2021resilient,norlab,cerberus}). Even
our team started off using RViz as a quick way to prototype
interfaces~\cite{Otsu2020IEEEAerospace} and used it as the main way to interact with
the robot agents due to its tight integration with ROS and ability to access
robot data for debugging purposes. We shifted away from this approach for the final competition, and the resulting HRI modalities and supervisory interface are presented in this work.

\section{Background and Objectives}
\label{sec:problem_requirements_objectives}
\ph{Challenge Requirements} The overall SubT goals are two common problems faced
by real-world multi-agent systems: first, the autonomous exploration of unknown
environments, and second, the search for objects of interest hidden within. While
exploration and search provide a need for specific capabilities, DARPA further
introduced a set of guidelines and rules to motivate higher levels of autonomy
for the deployed systems:
\begin{inparaenum}[(i)]
    \item only a single human operator is allowed to interact, supervise, and interface with the robots;
    \item each mission is bound by a fixed \textit{setup time} limit of 30 minutes and an \textit{exploration time} limit between 30 and 60 minutes;
    \item a pit crew of four (Finals) or nine (Urban Circuit) can support the supervisor by setting up hardware in a designated area without access to wireless data streams, robot control, or interface;
    \item there is a limited number of attempts to submit discovered objects of interest;
    \item the final challenge environment comprises tunnel, urban, and cave terrains to be explored.  
\end{inparaenum} 

\ph{Objectives} Deploying and operating large teams of robots like Team CoSTAR's robot fleet, shown in  \Cref{fig:robot_team}B, are complex real-world problems. Addressing this set of problems creates the need for a resource-efficient and robust human and multi-agent system to i) not overwhelm the single human supervisor, ii) meet the timing requirements, and iii) increase the performance of both exploration and search tasks.

To tackle this challenge and develop a system that can deploy reliably even beyond the SubT challenge, we embed the following interaction objectives into our system design:
\begin{inparaenum}[(1)]
    \item Reducing overhead and human workload (e.g., from application switching and manual task execution)
    \item Creating and maintaining situational awareness
    \item Managing large teams of robots (from setup, deployment to exploration) while allowing for a flexible configuration
    \item Accessing critical information in a single unified interface
    \item Maintaining an enjoyable performance that can visualize the complete robot team
    \item Collaborating with autonomy and trusting automation.
\end{inparaenum}


\section{Supervised Autonomy}
\label{sec:supervised_autonomy}
\subsection{Copilot}
\ph{Motivation} After SubT's ``Urban Circuit'', the allowed personnel in the
competition staging area was reduced from ten to five team members which
includes the main supervisor. This required a shift in how robots were
strategically and physically handled (minimum 2 people are needed to lift and
stage a single robot). Task coordination was done by a pit crew member directing
the operator and influencing their actions while following static paper
\textit{checklist procedures}. Developing and deploying a computerized assistant
that could take over this role was soon desired.

\ph{Original Implementation} A first version of Copilot, ``an autonomous
assistant for human-in-the-loop multi-robot operations'' was introduced in
\cite{Kaufmann2020copilotMike}. This early Copilot was only tested in realistic
cave simulations or during preparatory missions with one deployed robot. Copilot supports a single human supervisor in monitoring robot teams, aids with strategic task planning, scheduling, and execution, and communicates high-level commands between agents and a human supervisor if a communication link exists. The autonomy assistant aims at keeping workload acceptable while maintaining high situational awareness that allows rapid responses in case system failures are observed.
  
\ph{Task Interaction} Copilot takes over the decision-making processes regarding planning and scheduling, which reduces the need to memorize tasks and task sequences or the need to delegate a team member to take over such checklist-like tasks. Some tasks were implemented with higher autonomy levels and automatically executed limited actions, but most required the human to start the task, manually execute parts of it, and confirm that the task had been completed successfully or unsuccessfully while monitoring the system. 
On one hand, it reduced the need to remember tasks; on the other hand, more interactions with the newly introduced system were needed.

\ph{Scalability Limitations} Due to computational limitations, a full mission
simulation could not be achieved with more than three robots at reduced
real-time and not more than two in real-time. However, upon tightly integrating
Copilot with multiple real robot platforms, we noticed that the current concept
of operations didn't scale well when adding more robots to a mission. We learned
that task execution on the real hardware requires different timing and
introduces many sources for machine and human errors (e.g., if cables are loose,
sensors don't power up, or unknown unknowns occur).

\ph{Visualization Limitations} In robotics interfaces, scheduling, and timeline
views are often presented in a robot- or task-centric way, focusing on who or which
agent is scheduled for a certain task and when, respectively
\cite{BaeRossiDavidoff2020VizAnalytics}. The main task-centric approach that was
used in early Copilot tests showed a vertical list view with a scrollable
timeline. This timeline showed the four tasks closest in time on top. As the
number of tasks scaled linearly with the number of deployed robots this list
view became inefficient --- especially when tasks had to be deferred and worked
on in a non-sequential order.

\subsection{Improved Copilot}

The identified shortcomings motivated a redesigning and rethinking of Copilot's back-end and front-end to reduce and not just shift workload; thus, we implemented higher levels of automation.

\begin{figure}[t]
    \centering
    \def\svgwidth{\columnwidth}
    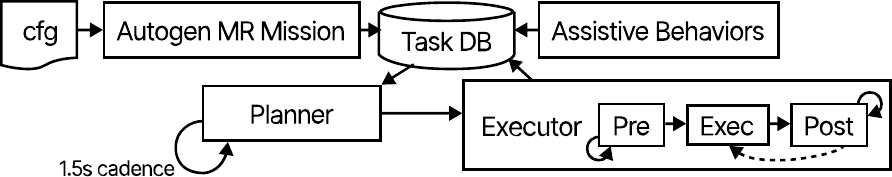
\caption{Copilot's task management architecture. Auto-generator, Planner, and Executor have been added or updated and access a centralized task database which stores pending, active, successful, or failed tasks.}
\label{fig:copilot_architecture}
\end{figure}

\ph{Architecture Changes} \Cref{fig:copilot_architecture} provides a simplified
overview of Copilot's updated task management architecture. A multi-robot task
auto-generator and verifiable task executor have been added to the system, and
the underlying planner has been replaced. All modules access a centralized task
database which stores pending, active, successful, or failed mission tasks for
setup, deployment, and during exploration.

\begin{figure}[!tb]
    \centering
    \def\svgwidth{\columnwidth}
    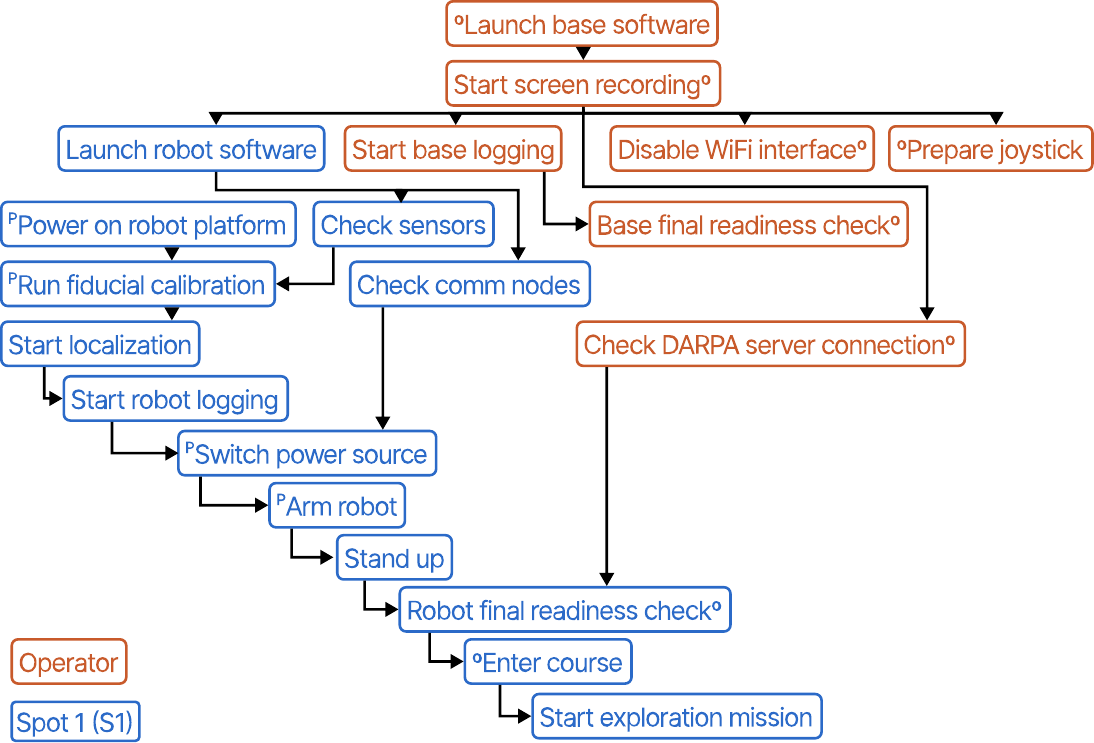
\caption{Pre-defined Copilot tasks for a single robot mission indicating task dependencies. The number of tasks scales linearly with the number of deployed robots. Spot1 related tasks are depicted in blue and operator tasks in orange. A superscript O or P at the beginning of a task indicate that the operator or pit crew has to manually fulfill some pre-condition. A superscript at the end indicates that a human sign-off is implemented before proceeding with the next task. For instance ``Power on robot platform'' requires a physical push of the robot platform startup button.}
\label{fig:task_dependencies}
\end{figure}

\ph{Task Dependency Graph} A robot mission can be fairly complex, even when
looking at the deployment of a single robot. In
\Cref{fig:task_dependencies} such a single robot mission is shown as a directed
graph indicating the temporal
constraints and execution dependencies with arcs between the nodes that represent a
pre-defined set of mission tasks. Each task is defined by its duration, earliest
start time, latest end time, and its dependency relations with other tasks.

To deploy multiple robots without the need to hard-coding all possible agent
combinations and graphs, we use a scalable auto generator. The preceding
superscript O in the graph (see \Cref{fig:task_dependencies}) indicates
that human inputs or actions are required for the task. In the case of the
\texttt{Launch base software} task, this means that the operator has to initiate
the software launch as a pre-condition and is prompted to select the robots that
they would like to deploy for the upcoming mission. Similarly, superscripts at
the end of a task indicate that human action is needed before the next task can
begin. Tasks without either have been fully automated for nominal cases in this
newer Copilot version.

\ph{Task Planning and Scheduling} The aforementioned task dependency graph for
the selected robots forms the input for Copilot's task planner and is stored in
the MongoDB task database. The generation of a task plan for setting up, deploying, and
assisting the operator during exploration is framed as an automated temporal
planning problem. In the first version of Copilot, we formulated such problem as
a Simple Temporal Network (STN), encoded as a linear program. In the improved
version of Copilot, deployed in the final events of SubT, we moved to a PDDL
temporal planning formulation to allow 1) flexibility on task representation
with respect to state constraints, resources, and planning, and 2) use the body
of planners available in the literature. Herein we integrated the OPTIC planner
\cite{benton-etat-2012-OPTIC}, a PDDL temporal planner that handles time window
specification (timed initial literals), and discrete and continuous resources.

To perform planning, OPTIC uses both a PDDL domain file and a problem
file. The domain file has been designed to represent tasks (modelled as
operators) and its dependencies (preconditions). The problem file is generated
prior to calling the planner, and it is built based on the current state of
mission and tasks execution. For example, if a task is ongoing, the PPDL file
would represent the task as ongoing and add constraints to ensure it continues
the execution to meet the necessary constraints. As a notional example
of the scale of the planning problem, a mission with four robots would have
approximately 60 tasks to be scheduled during setup and deployment. Planning is
performed at a predefined cadence (e.g., every 1.5 seconds), but it also follows
an event-based approach when task execution is late, or the human-in-the-loop
changes their strategy --- this helps mitigate execution uncertainty. The
generated plan is parsed and stored in a Task Database (for logging and
visualization across the system); each task is then dispatched for execution.

If a plan is not found by OPTIC due to temporal constraint violations (e.g.,
delays in task execution), Copilot will attempt to increasingly relax some of
the key temporal constraints, such as the latest end time of certain activities
(e.g., allowing setup tasks to end a few minutes after the setup time,
overlapping with the beginning of the exploration time window). In critical
scenarios, Copilot would notify the operator of a schedule relaxation to allow
for further strategy changes.

\ph{Task Verification and Execution} A verifiable and generic task framework is
introduced to Copilot, allowing for quick implementations and standardized task
automation. Each task follows a strict precondition, execution, and
post-condition template. Condition checks and execution can be triggered across
agents, including the base station at which the human can oversee all automated
processes at a high level in the new Copilot interface, which is described in
\Cref{sec:interface}. The task template execution covers both fully automated
tasks and semi-automated tasks where an operators confirmation is required (e.g.
deploying a robot into a cave requires a Go/No-go decision from the supervisor ---
deploying itself is an automated process). If a task fails during execution or
post-condition checking, Copilot will try to resolve the issue by retrying tasks
several times and allowing for more execution time. Failed tasks will be reported
to the supervisor, who can choose to debug the issue at hand or trigger another
automated retry. Retries and resets are possible at all levels, and completed
tasks can be reset during an active mission in case a robot platform has to be
rebooted.


\section{Game-Inspired Interface}
\label{sec:interface}
	
\begin{figure*}[tb]
\centering
\includegraphics[width=0.8\textwidth]{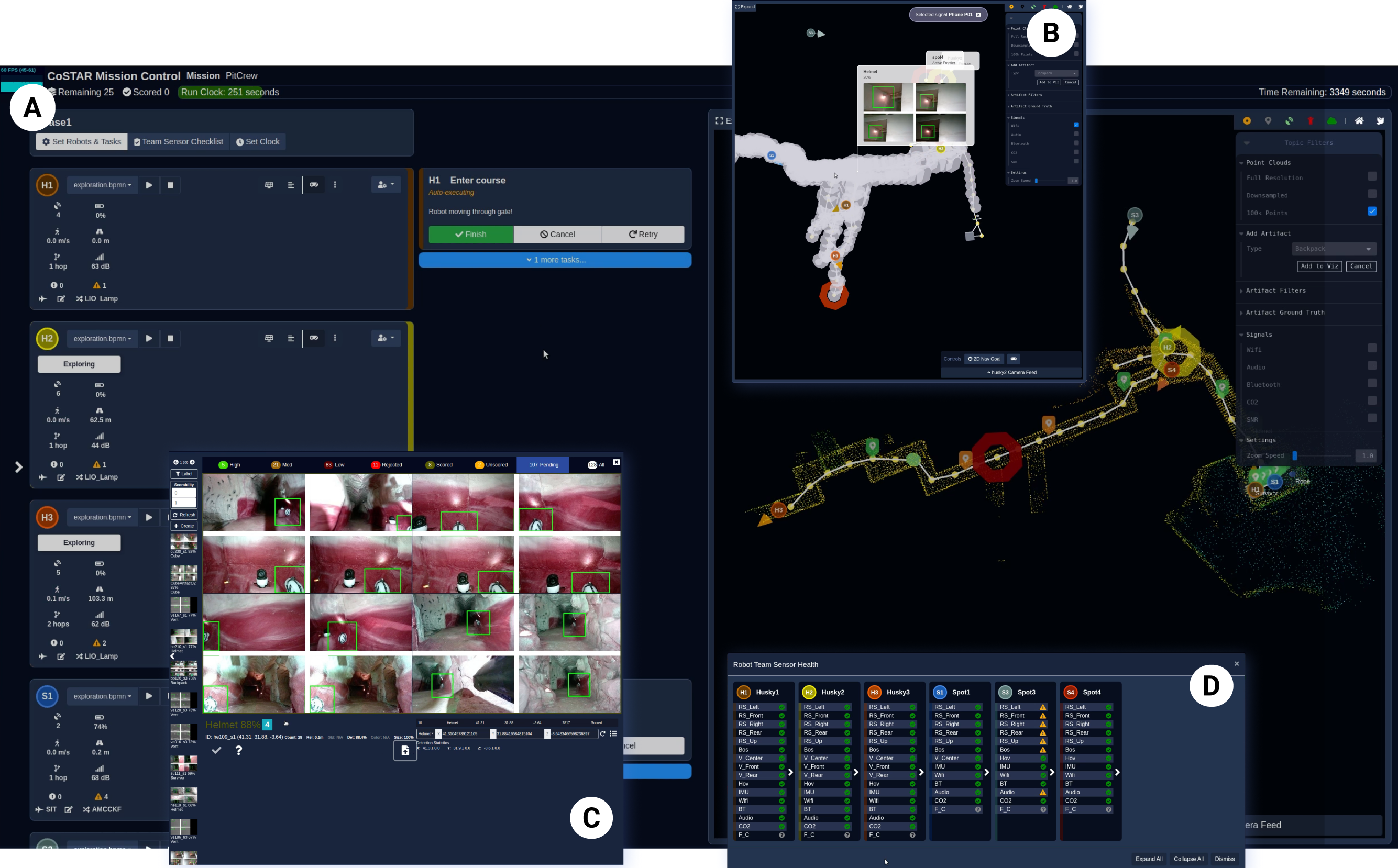}
\caption{An overview of the major UI components. (A) The Robot and associated Copilot task cards.  (B) The split-screen 3D visualization view with view controls, WiFi signal strength overlay, and an artifact card showing on the map. (C) The artifact drawer. (D) The robot health systems component.}
\label{fig:ui_overview}
\end{figure*}

\ph{Game Inspiration} Inspiration for multi-agent interaction and interface
design is partially drawn from real-time strategy games such as Age of Empires,
StarCraft, and {Command \& Conquer}. When played competitively, these games
require a high sense of micro and macro-management of units and their
environment and the ability to efficiently switch between these two ways of
managing a team. Micromanagement involves short-term strategy and
decision-making, where individual units may require critical attention to win a
battle, overcome an obstacle, or navigate to the next point of interest, while
macromanagement refers to longer-term strategizing that involves resource
gathering, unit production over time, and overall exploration and control of the
map~\cite{rtsCombatStrategy}. Parallels can be applied to the management of a
robot team in the SubT competition. Even autonomous robots can benefit from or
require human intervention and commanding, especially if critical attention
towards failing subsystems is needed. Supervised multi-agent control draws from
the human's situational awareness regarding the environment and robot states to
effectively coordinate multi-agent behaviors, successfully locate artifacts, and
score points.

\ph{Mission Phases} The user interface is designed to be adaptable to the
overall mission and two major phases of an individual robot's competition run in
particular:
1) setup and deployment, and 2) mission execution with its exploration and
search components. Across these phases, the visibility and abstraction of
information need to be flexible to facilitate focus on the anticipated operator
interactions. In deployment, the user interface uses the Copilot-generated tasks and status information to guide the sole operator through the multitude of individual tasks while allowing them to maintain their situational awareness, manage the entire robot team, and coordinate with the pit crew.

\ph{Three Column Layout} The Mission Control interface is organized into
different view components. \Cref{fig:ui_overview} shows the main split-screen
with three columns aiming at creating reliable locations for the operator to
look at when needing to accomplish functionally distinct tasks (A). The aim here
is to reduce the amount of visual scanning, application switching, and to parse
robot needs on an individual or team level swiftly. Individual robot information
pertinent to monitoring health systems is available on the left, planned and
actively re-scheduled Copilot tasks for individual robots are placed alongside
each agent in the middle, and a 3D interactive visualization of the robots in
their environment is anchored to the right. During mission execution, the
primary goal of the user interface is to keep the operator situationally aware
of a multitude of individual robot health systems and data sensed from the
surrounding environment while presenting the most important information and thus
reducing their cognitive workload. In \Cref{fig:robot_team} the 3D visualization
is expanded, and robot sensor and status information is minimized to select
mission-critical information.
    

\ph{Health Systems and Robot Status} In order to effectively survey the status
of any individual robot in the team, visibility into over 30 unique sensors and
statuses needed to be surfaced to the operator per robot. This required
identifying which indicators were critical to display at all times, which could
be hidden within a sub-view, which were good candidates to be combined and
abstracted, and which would be prioritized across either the deployment (split)
or mission execution (split and expanded visualization) modes of the user
interface. In addition to sensors visible at an individual level, an additional
view was created to organize sensors compactly across the team, providing easy
visual scanning for the operator during macro-management and deployment, as
shown in \Cref{fig:ui_overview}D. An abstraction of robot behaviors (e.g.,
exploring, dropping a communications node) and mobility states presents an
overall status of each robot to the operator by color and a high-level
description. This status is prioritized based on criticality to ensure the
operator's attention will be requested for the most important issue at any given
time.

Previously, Copilot tasks resided in an entirely separate module of the
interface with limited screen estate, requiring the operator to move other
related and necessary sensor and status information out of physical view. A
reorganization where Copilot tasks are paired alongside their respective robots
is utilized to reduce context loss and pair necessary information to complete
the tasks together, as shown in \Cref{fig:ui_overview}A. Over time during the
development roll-out, this pairing of health, sensor, and status indicators
alongside Copilot tasks facilitated a level of trust from the operator where
focus on a particular robot was not necessary unless a critical task requiring
operator intervention appeared.

\ph{3D Visualization View} A 3D interactive visualization leveraging React Three
Fiber (a React-based renderer for three.js) was created within the UI with the
aim of achieving a significant reduction in operator task and application
switching. Prior to this version of the interface, the operator was required to
switch between a web browser to view robot health systems and status information
and RViz (a visualizer for ROS) to view the robots within the 3D environment and
command them. In the split view of the UI, the operator can have the full
context of robot sensors and status information along with any outstanding
Copilot tasks. When in the expanded visualization view, the layout shares
similarities with layouts of traditional Real-Time Strategy (RTS) games, where
content is functionally organized from the corners of the view and leave the
center-most screen real estate where the operator will primarily interact with
robots and information unobstructedly. From this view, the operator can take
on any of the following tasks: surveying the mapped environment and robot
positions for locations to scout, locating, and submitting object or signal
artifacts, directing or course-correcting robot autonomy with manual navigation
commands, viewing signal strength of the communications backbone within the
environment, and assigning robots to drop communication nodes manually. The
visualization allows the operator to navigate the 3D environment through
panning, zooming, and filtering points of interest categories. To effectively
manage the switching between micro and macro-level interactions, a single-click
shortcut was implemented on each robot status card for the operator to quickly
focus on any robot that requires attention. An additional shortcut is
provided to zoom back out to an overview of the map.
	
Improvements over traditional RTS commanding controls were also made to minimize
the amount of mouse control and coordination necessary. Instead of requiring to
select or drag a bounding box prior to commanding a robot, the operator could
simply interact with the visualized information roadmap (IRM) --- a breadcrumb
trail used for safely navigating the environment constructed by the team of
robots~\cite{plgrim} --- and assign any robot with a high priority navigation point or
communications node drop location through a context menu, regardless of whether the
particular robots are currently in view or not.
	
To help with artifact management, the locations of detected artifacts are
visualized and interactivity is added to allow the operator to quickly hover
into a thumbnail and click to navigate to the dedicated Artifact Drawer
\Cref{fig:ui_overview}C for deeper analysis and submission. Additional
interactions are, for example, manually adding and manipulating detected artifact
locations within the 3D space, by dragging its location across a plane for
fine-tuning if a submission location was deemed incorrect and needed adjustment.
	
While in the expanded visualization view, compressed versions of the robot
status modules are shown horizontally in the bottom left of the view with the
mission status indicator made more prominent and placed above each module. These
overall status indicators were given visual priority to ensure grabbing the
operator's attention. For instance, the indicator would flash red when a robot
had fallen over, was low on battery, or required assistance. The operator could
immediately click the respective robot module and be oriented over it for
micromanagement.

\ph{Artifact Drawer} Artifact submission was a critical part of SubT that also has many real-world parallels, for instance, in search and rescue. Especially under time constraints, it is necessary to quickly identify artifacts of interest in the environment, whether these be human survivors or other objects of interest. Detecting and localizing artifacts automatically is done using a state-of-the-art image processing pipeline \cite{terry2020object}, but no AI system is infallible, especially in unknown environments, so having a system for an operator to manually review artifacts efficiently was critical considering mission time and submission attempts.

In the old system \cite{terry2020object}, a manual artifact review system did exist, but it was built with a focus on only basic functionality and a high reliance on initially accurate artifact detections. Each artifact report took roughly 90 seconds to review. In redesigning this component, we wanted to focus on improving the review process from an ease of use perspective and decrease the time spent to confidently review an artifact report down to 15 seconds. Beyond simply making the system more intuitive for the operator, this actually had a major functional benefit from a trustability standpoint in that it allowed us to decrease the confidence threshold for flagging artifact detections and have the operator go through and verify nearly 6 times more potential artifact reports while not increasing total time spent. 
    
To better design the new system for speed, it was important to understand which areas of the old one were slowing the process down the most. Testing the old system in simulation and operator feedback revealed that the artifact review process needed too many clicks. Then, time had to be spent zooming in on and reviewing images and checking with RViz separately to verify that artifact coordinates were correct. No visual aid was given if corrections were necessary, and coordinates had to be updated by manually entering them for each axis in ${\rm I\!R^3}$. Borrowing from game interface design, integrating the 3D visualization view directly into the web UI removed the need for application switching, and drag controls were added to adjust locations providing correctly scaled coordinate updates from the 3D environment. A minified list that provides an overview of all artifact reports by confidence levels, plus maximizing the screen real estate of a single selected artifact helped increase efficiency. Finally, adding keyboard shortcuts as commonly used in gaming made meeting our target goal of 15 seconds possible. 


\section{Results}
\label{sec:results}

Over the course of the last challenge year, we conducted a limited series of field tests in three testing locations, including the abandoned tunnels at the Los Angeles Subway Terminal building, the Lava Bed National Monument in Northern California, and the Kentucky Underground lime-stone cave for which we applied our rapid development and testing strategy. We experimented with different robot configurations and in different stages of readiness as our system's capabilities matured. We deployed up to 11 vehicles simultaneously during these tests stressing the overall system (including Copilot and all the UI elements) and learning about its technical limitations like bandwidth and computing resources which will be presented in upcoming work.

We deployed the presented game-inspired user interface and supervised autonomy system during the SubT challenge using four to six ground robots nominally. While we could have exceeded the number of six robots using the newly designed interface and autonomy, six became the preferred number of agents to explore large-scale environments while allowing reliable communication links that would not exceed bandwidth limitations when robots disseminated information from autonomously explored out-of-comms areas. This allowed meeting the set interaction objectives, especially maintaining an enjoyable performance that can visualize the complete robot team while contributing to a lower workload due to fewer deployed agents.

In what follows, we analyzed screen recordings and log files (approved by Caltech IRB protocol number 19-0461 and Polytechnique Montreal project CER-2122-50-D) collected during the SubT final competition. We extracted time-to-task information, robot deployment times, mouse locations, and application usage from runs P1, P2, and F that consist of a setup-time and mission phase of 30+30 and 30+60 minutes, respectively. Robots were only allowed to leave the setup area and enter the course when the mission time began. Readying the team of robots and not bleeding into the mission time was a crucial effort to maximize available mission and exploration time. The results are compared to an earlier state of the system that did not implement Copilot and used different interfaces, namely the SubT ``Urban Circuit'' similar to \cite{Otsu2020IEEEAerospace}. During the ``Urban Circuit'' task, coordination was done by humans only.

\begin{figure}[t]
\centering
  \includegraphics[width=0.5\textwidth]{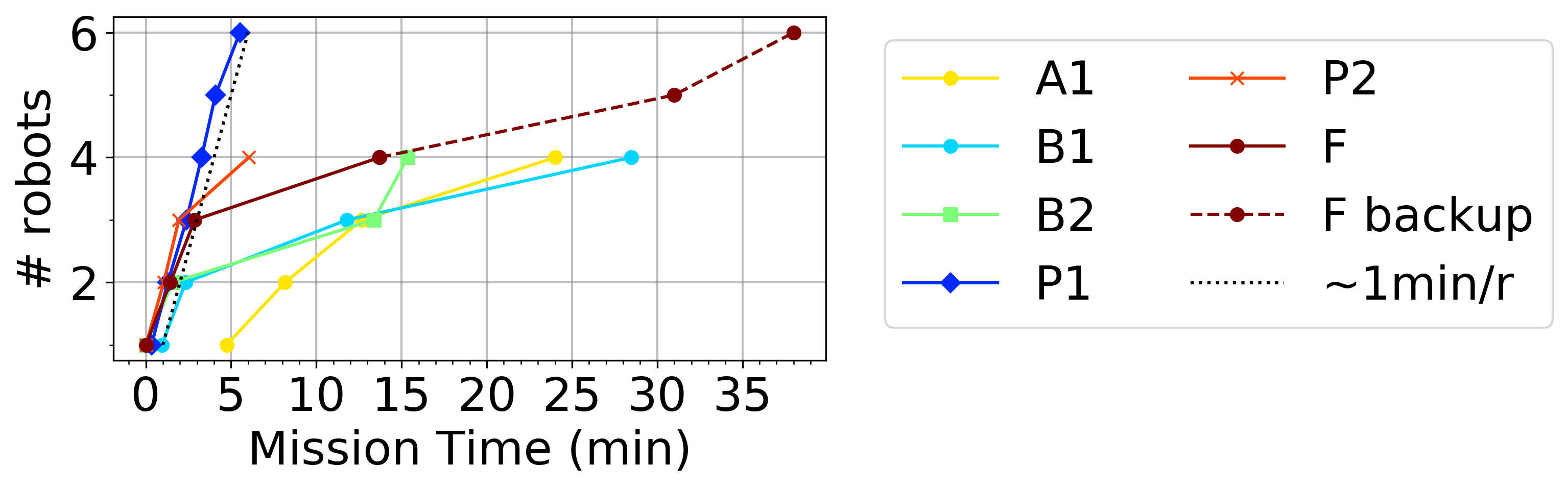}
  \caption{Robot deployment times per game run measured upon entering the course. The black dotted line ($\sim$1min/r) indicates the team's internal goal for robot deployment and represents a deployment of one robot per minute. F backup marks insertion points of 2 robots that were not part of the initial deployment strategy but were added ad-hoc to compensate for robot failures during run F.}
  \label{fig:robot_deployment}
\end{figure}

\ph{Robot Deployment} \Cref{fig:robot_deployment} shows the robot deployment times that were achieved by deploying Copilot and compares them to the baseline. We can see that during run P1, we achieved sending one robot in less than 60 seconds each, deploying a total of 6 ground vehicles in 5 minutes and 31 seconds. In runs P2 and F, we achieved staying below the one minute mark for the first three robots. Deploying the robots without Copilot and the new interface in the `Urban Circuit'' runs A1, B1, and B2 took more than 5.5 min per robot on average, thus significantly reducing the time available for exploration and consequently reducing ground coverage and information gain regarding the search task. 

\begin{figure}[t]
  \centering
  \includegraphics[width=0.5\textwidth]{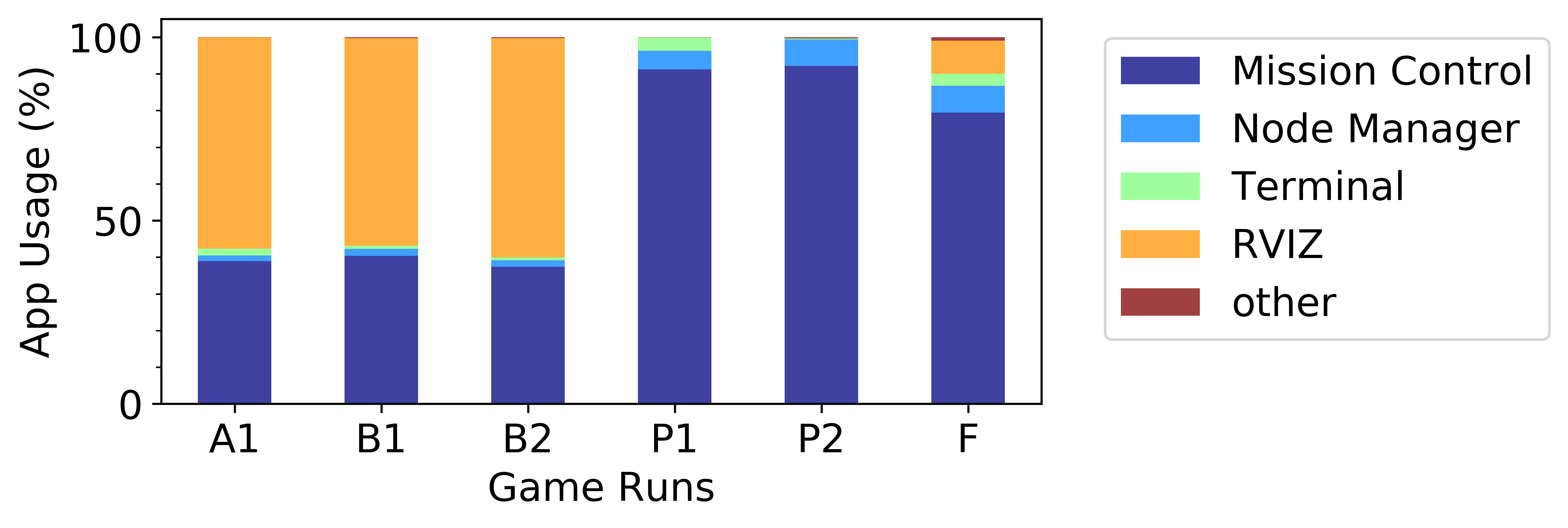}
  \caption{Application usage (foreground application) for six SubT mission runs in percent. A1, B1, and B2 represent the usage before the redesign that integrated 3D visualization and interactions for P1, P2, and F in a single Mission Control application using only one computer and screen. Note that node manager and terminal usage are underrepresented in runs A1, B1, and B2 because the initial setup phase of up to 10 minutes was not recorded for these runs due to different logging procedures.}
  \label{fig:application_usage}
\end{figure}

\ph{Application Usage} The new interface resulted in a shift in application usage and reduced switching between different applications and computers with a second set of peripherals, as RViz was running on a second device during the ``Urban Circuit''. \Cref{fig:application_usage} presents the relative usage of applications for six SubT runs. Designing a unified interface resulted in a shift in application usage that reduced the use of RViz significantly. While more than 50\% of time was spent on RViz during the ``Urban Circuit'' runs, we were able to unify user interactions and situation awareness in a single Mission Control interface. Only run F uses RViz for some time as a debugging tool that gave access to the robot's cost maps depicting the perceived risks around them. This information was not visualized by the new interface, but presents valuable key information in case of unexpected and off-nominal operations.

\begin{figure}[t]
  \centering
  \includegraphics[width=0.23\textwidth]{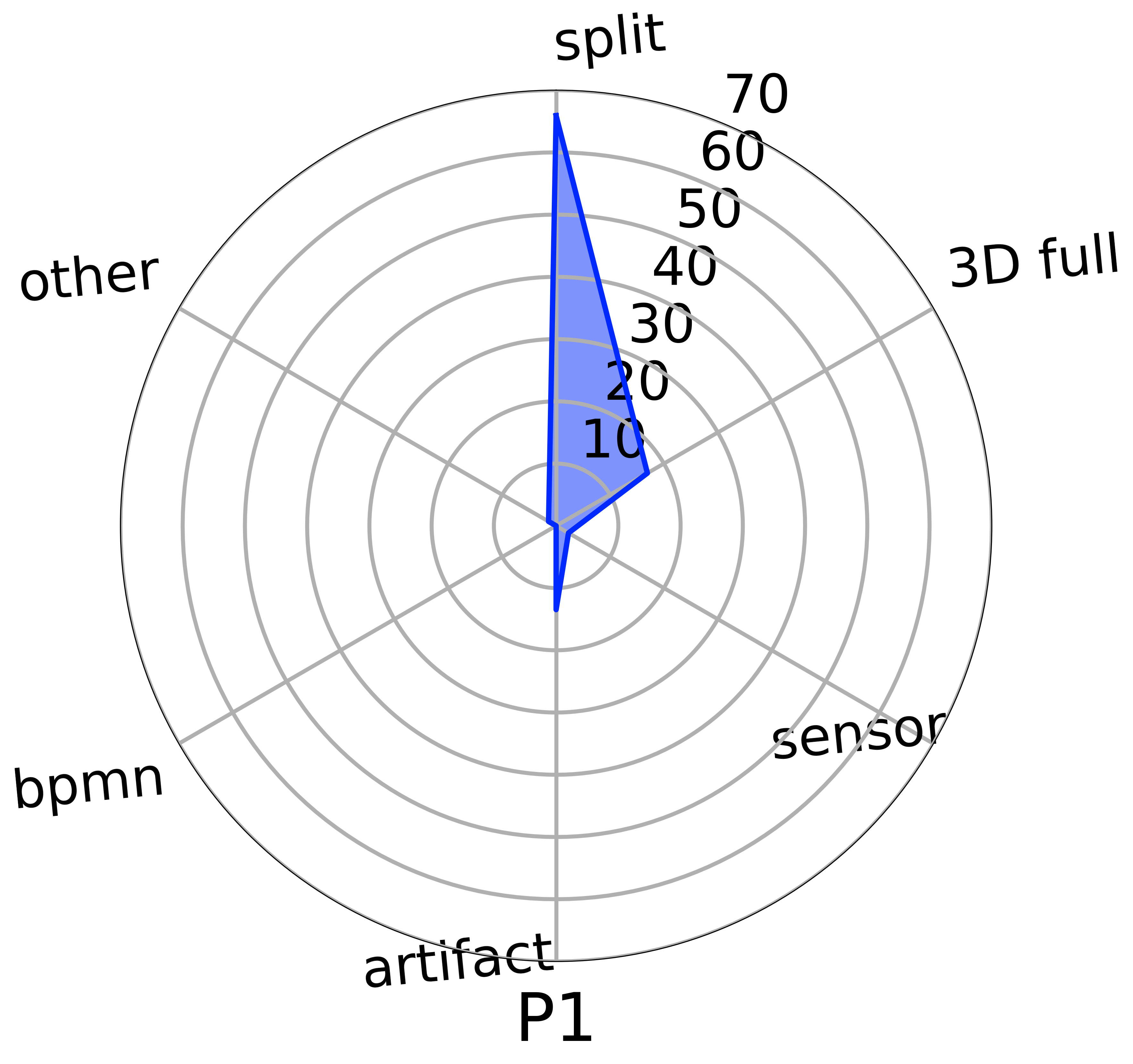}
  \includegraphics[width=0.23\textwidth]{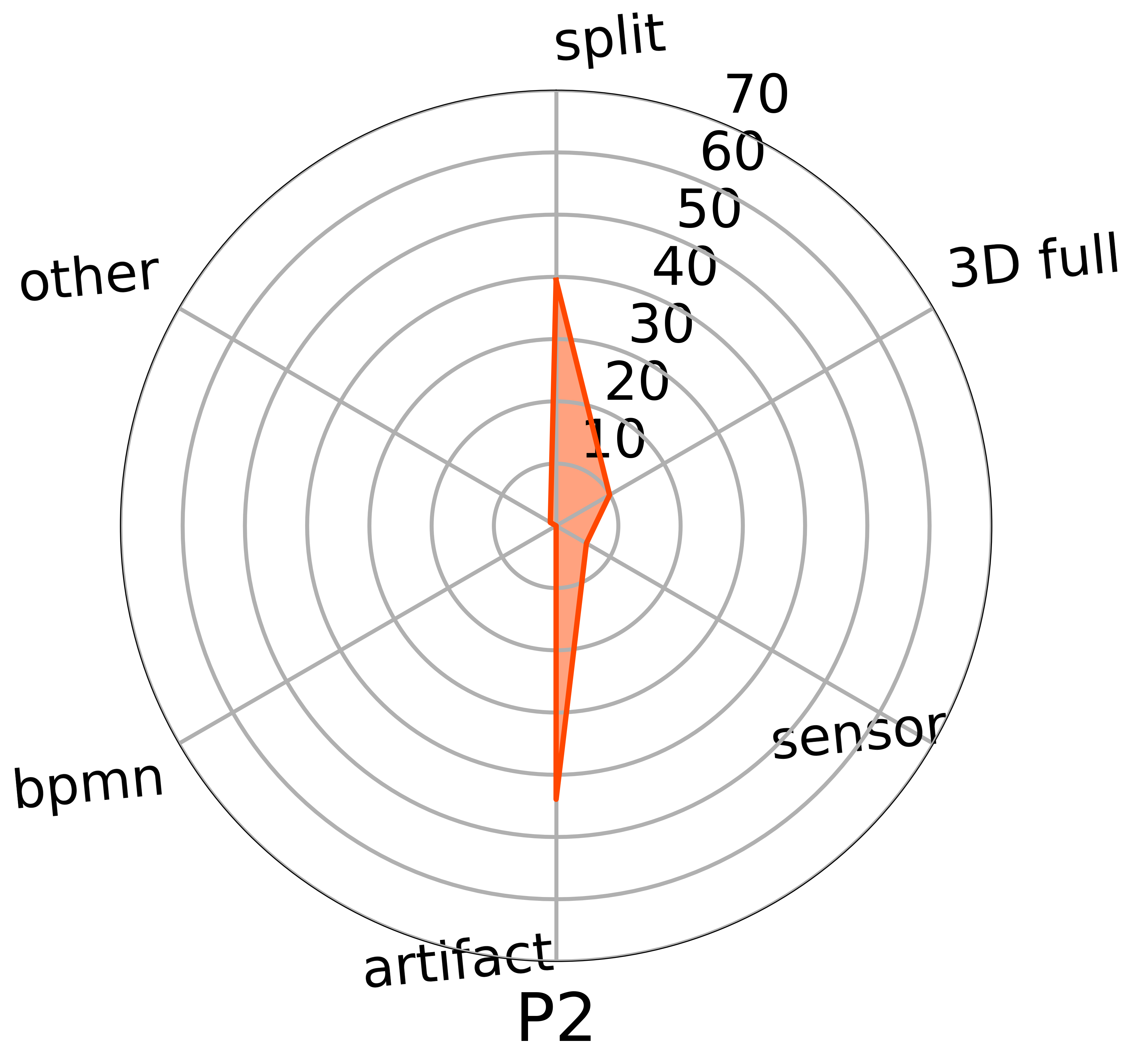}
  \includegraphics[width=0.27\textwidth]{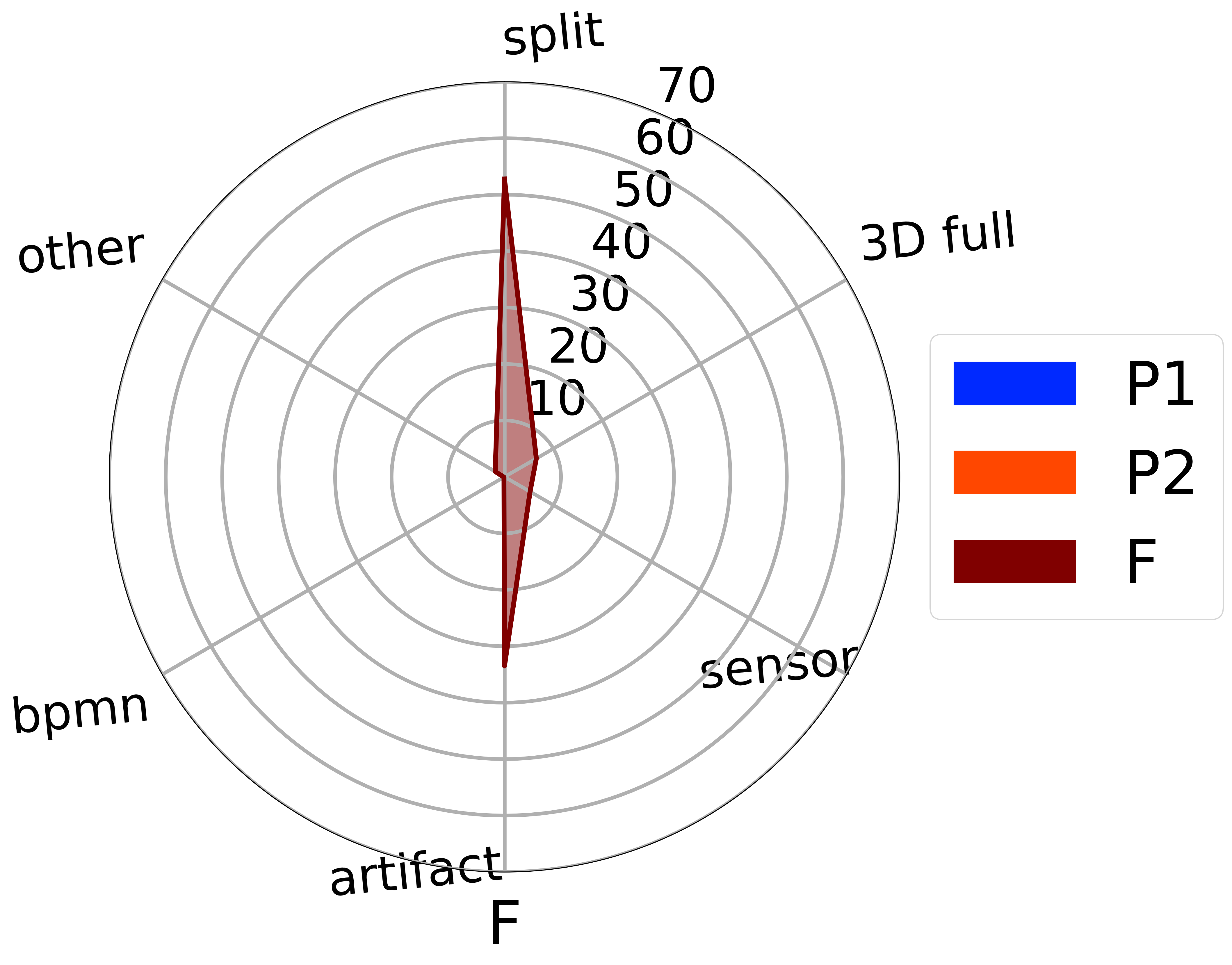}
  \caption{Analysis of the redesigned user interface interaction by view component in percent for runs P1, P2, and F.}
  \label{fig:radar_plot}
\end{figure}

\begin{figure}[t]
  \centering
  \includegraphics[width=0.5\textwidth]{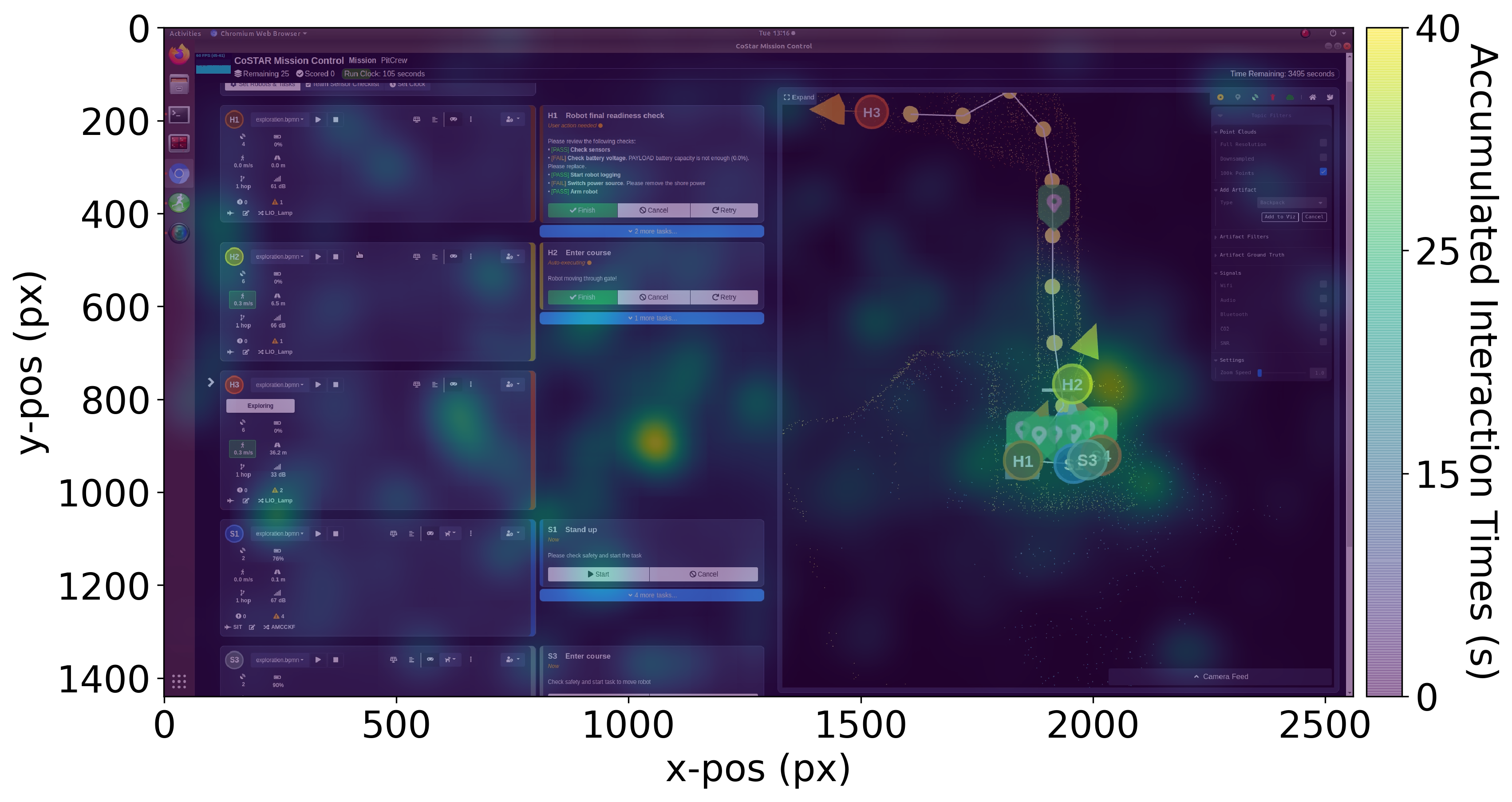}
  \caption{Activity heat map showing the x and y positions of cursor interactions (and indirectly gaze) overlaid on the Mission Control Split-Screen view exemplary for game run P2. The view consists of robot cards, a column for Copilot tasks, and the split-screen 3D view. A brighter heat map indicates higher interaction times in this area. Stationary cursors for more than 10 seconds are classified as inactive.}
  \label{fig:heatmap}
\end{figure}

\ph{UI Feature Usage} With the main Mission Control interface being the main
interaction point for human supervisory control, we then look at the feature
usage within the interface itself. \Cref{fig:radar_plot} shows the relative
interaction times with the split-screen view, the 3D full-screen console view,
the sensor health overview, artifact submission drawer, and the BPMN modal that
gives a detailed overview of a robot's inner state machine (which was relied
upon during the ``Urban Circuit''). We see that, especially during runs P2 and F,
large amounts of time were spent on the artifact drawer and thus performing the
search task analyzing the artifact reports that were generated by the
multi-agent system. To gain situational awareness and potentially interact with
the robot team, the human supervisor primarily relied on the split-screen view of
the Mission Control app that is shown in the background of \Cref{fig:heatmap}
overlaid by a heat map that indicates the most active areas derived from mouse
cursor positions sampled at 1.5 Hz. In this analysis, an area is deemed inactive
if the mouse has been stationary for more than ten seconds. \emph{Huang et
  al.}~\cite{andwhite2012user} found that the median difference between human
gaze and mouse position during an active task is 77 pixels with a standard
deviation of 33.9 pixels at 96 dpi screen resolution. A Gaussian kernel with
$\mu=98$ and $\sigma=43$ adjusting for 122 dpi is used to derive our heat maps.
\Cref{fig:heatmap} indicates that the robot cards, Copilot tasks and the 3D view
were all crucial tools while overseeing the robotic system and performing the
exploration and search tasks.
   
\section{Conclusions and Future Work}
\label{sec:conclusions_future_work}
In this work we 
\begin{inparaenum}[(i)]
    \item create a game-inspired user interface for multi-agent robot missions
    \item integrate an automated planner for task planning and scheduling,
    \item add a verifiable task framework for increased reliability, and
    \item present results on how the overall system performed over the course of
      several real-world deployments, including the DARPA SubT Challenge final.
\end{inparaenum}
In future work, we plan to deploy our interface and Copilot during scientific
exploration missions to autonomously map and identify geological features and
assess exploration strategies in lava tubes. This will lead to further
validation of the subsystems and a structured assessment of a supervisor's
workload outside the realm of the SubT challenge with experts and potentially
non-expert users. Ultimately, we would like to assess operator workload from
wearable sensors in real-time and consider such constraints in Copilot's task
planning.


\section*{Acknowledgment}
\footnotesize{The work is partially supported by the Jet Propulsion Laboratory, California Institute of Technology, under a contract with the National Aeronautics and Space Administration (80NM0018D0004), and Defense Advanced Research Projects Agency (DARPA). This work was conducted in collaboration with the Making Innovative Space Technologies Laboratory (MIST Lab) at Polytechnique Montreal. The first author would like to thank the Natural Sciences and Engineering Research Council of Canada (NSERC) for their generous support in the form of a Vanier Canada Graduate Scholarship. Thank you to all members of Team CoSTAR for their valuable discussions and support.}

\bibliographystyle{IEEEtran}
\bibliography{root}

\begin{thebibliography}{10}
\providecommand{\url}[1]{#1}
\csname url@rmstyle\endcsname
\providecommand{\newblock}{\relax}
\providecommand{\bibinfo}[2]{#2}
\providecommand\BIBentrySTDinterwordspacing{\spaceskip=0pt\relax}
\providecommand\BIBentryALTinterwordstretchfactor{4}
\providecommand\BIBentryALTinterwordspacing{\spaceskip=\fontdimen2\font plus
\BIBentryALTinterwordstretchfactor\fontdimen3\font minus
  \fontdimen4\font\relax}
\providecommand\BIBforeignlanguage[2]{{%
\expandafter\ifx\csname l@#1\endcsname\relax
\typeout{** WARNING: IEEEtran.bst: No hyphenation pattern has been}%
\typeout{** loaded for the language `#1'. Using the pattern for}%
\typeout{** the default language instead.}%
\else
\language=\csname l@#1\endcsname
\fi
#2}}

\bibitem{asada2019robocup}
M.~Asada, P.~Stone, \emph{et~al.}, ``{RoboCup}: A treasure trove of rich
  diversity for research issues and interdisciplinary connections {[TC
  Spotlight]},'' \emph{IEEE Robotics Automation Magazine}, vol.~26, no.~3, pp.
  99--102, 2019.

\bibitem{Hambuchen2017NASA_Rob_Challenge}
\BIBentryALTinterwordspacing
K.~A. Hambuchen, M.~C. Roman, \emph{et~al.}, \emph{NASA's Space Robotics
  Challenge: Advancing Robotics for Future Exploration Missions}.\hskip 1em
  plus 0.5em minus 0.4em\relax American Institute of Aeronautics and
  Astronautics, 2017. [Online]. Available:
  \url{https://arc.aiaa.org/doi/abs/10.2514/6.2017-5120}
\BIBentrySTDinterwordspacing

\bibitem{Link2021_ESA_ESRIC}
\BIBentryALTinterwordspacing
M.~Link and B.~Lamboray, \emph{European Space Resources Innovation Centre –
  ESRIC}.\hskip 1em plus 0.5em minus 0.4em\relax American Institute of
  Aeronautics and Astronautics, 2021. [Online]. Available:
  \url{https://arc.aiaa.org/doi/abs/10.2514/6.2021-4012}
\BIBentrySTDinterwordspacing

\bibitem{agha2021nebula}
\BIBentryALTinterwordspacing
A.~Agha, K.~Otsu, \emph{et~al.}, ``Nebula: Quest for robotic autonomy in
  challenging environments; {TEAM} costar at the {DARPA} subterranean
  challenge,'' \emph{Submitted to the Journal of Field Robotics}, 2021.
  [Online]. Available: \url{https://arxiv.org/abs/2103.11470}
\BIBentrySTDinterwordspacing

\bibitem{Titus2021}
\BIBentryALTinterwordspacing
T.~N. Titus, J.~J. Wynne, \emph{et~al.}, ``A roadmap for planetary caves
  science and exploration,'' \emph{Nature Astronomy}, vol.~5, no.~6, pp.
  524---525, 2021. [Online]. Available:
  \url{https://doi.org/10.1038/s41550-021-01385-1}
\BIBentrySTDinterwordspacing

\bibitem{Licklider1960ManComputerSymbiosis}
J.~C.~R. Licklider, ``Man-computer symbiosis,'' \emph{IRE Transactions on Human
  Factors in Electronics}, vol. HFE-1, no.~1, pp. 4--11, 1960.

\bibitem{chen2021human}
J.~Y. Chen and M.~J. Barnes, ``Human--robot interaction,'' \emph{Handbook of
  human factors and ergonomics}, pp. 1121--1142, 2021.

\bibitem{Kaufmann2020copilotMike}
M.~Kaufmann, T.~S. Vaquero, \emph{et~al.}, ``Copilot mike: An autonomous
  assistant for multi-robot operations in cave exploration,'' in \emph{IEEE
  Aerospace Conference}, 2021.

\bibitem{kohlbrecher2015human}
S.~Kohlbrecher, A.~Romay, \emph{et~al.}, ``Human-robot teaming for rescue
  missions: Team vigir's approach to the 2013 darpa robotics challenge
  trials,'' \emph{Journal of Field Robotics}, vol.~32, no.~3, pp. 352--377,
  2015.

\bibitem{cerberus}
M.~Tranzatto, F.~Mascarich, \emph{et~al.}, ``\BIBforeignlanguage{en}{Cerberus:
  Autonomous legged and aerial robotic exploration in the tunnel and urban
  circuits of the darpa subterranean challenge},''
  \emph{\BIBforeignlanguage{en}{Field Robotics}}, 2022.

\bibitem{Szafir2021HRI_DATA_VIZ}
\BIBentryALTinterwordspacing
D.~Szafir and D.~A. Szafir, ``Connecting human-robot interaction and data
  visualization,'' in \emph{Proceedings of the 2021 ACM/IEEE International
  Conference on Human-Robot Interaction}.\hskip 1em plus 0.5em minus
  0.4em\relax New York, NY, USA: Association for Computing Machinery, 2021, p.
  281–292. [Online]. Available: \url{https://doi.org/10.1145/3434073.3444683}
\BIBentrySTDinterwordspacing

\bibitem{rahmani2019space}
A.~Rahmani, S.~Bandyopadhyay, \emph{et~al.}, ``Space vehicle swarm exploration
  missions: A study of key enabling technologies and gaps,'' \emph{Proceedings
  of the 70th International Astronautical Congress}, 2019.

\bibitem{coactive_design}
\BIBentryALTinterwordspacing
M.~Johnson, J.~M. Bradshaw, \emph{et~al.}, ``Coactive design: Designing support
  for interdependence in joint activity,'' \emph{J. Hum.-Robot Interact.},
  vol.~3, no.~1, p. 43–69, feb 2014. [Online]. Available:
  \url{https://doi.org/10.5898/JHRI.3.1.Johnson}
\BIBentrySTDinterwordspacing

\bibitem{Roundtree2019VizDesignCollectiveTeams}
\BIBentryALTinterwordspacing
K.~A. Roundtree, J.~R. Cody, \emph{et~al.}, ``Visualization design for
  human-collective teams,'' \emph{Proceedings of the Human Factors and
  Ergonomics Society Annual Meeting}, vol.~63, no.~1, pp. 417--421, 2019.
  [Online]. Available: \url{https://doi.org/10.1177/1071181319631028}
\BIBentrySTDinterwordspacing

\bibitem{Wallner2019Playtesting}
\BIBentryALTinterwordspacing
G.~Wallner, N.~Halabi, and P.~Mirza-Babaei, ``Aggregated visualization of
  playtesting data,'' in \emph{Proceedings of the 2019 CHI Conference on Human
  Factors in Computing Systems (CHI)}.\hskip 1em plus 0.5em minus 0.4em\relax
  New York, NY, USA: Association for Computing Machinery, 2019. [Online].
  Available: \url{https://doi.org/10.1145/3290605.3300593}
\BIBentrySTDinterwordspacing

\bibitem{medlock2002usingRITE_AgeOfEmpires}
M.~C. Medlock, D.~Wixon, \emph{et~al.}, ``Using the {RITE} method to improve
  products: A definition and a case study,'' \emph{Usability Professionals
  Association}, vol.~51, 2002.

\bibitem{csiro}
\BIBentryALTinterwordspacing
N.~Hudson, F.~Talbot, \emph{et~al.}, ``Heterogeneous ground and air platforms,
  homogeneous sensing: Team {CSIRO} data61's approach to the {DARPA}
  subterranean challenge,'' \emph{CoRR}, 2021. [Online]. Available:
  \url{https://arxiv.org/abs/2104.09053}
\BIBentrySTDinterwordspacing

\bibitem{cornellSubTJFR}
\BIBentryALTinterwordspacing
M.~T. Ohradzansky, E.~R. Rush, \emph{et~al.}, ``Multi-agent autonomy:
  Advancements and challenges in subterranean exploration,'' \emph{CoRR}, 2021.
  [Online]. Available: \url{https://arxiv.org/abs/2110.04390}
\BIBentrySTDinterwordspacing

\bibitem{scherer2021resilient}
S.~Scherer, V.~Agrawal, \emph{et~al.}, ``Resilient and modular subterranean
  exploration with a team of roving and flying robots,'' \emph{Submitted to the
  Journal of Field Robotics}, 2021.

\bibitem{norlab}
\BIBentryALTinterwordspacing
T.~Roucek, M.~Pecka, \emph{et~al.}, ``System for multi-robotic exploration of
  underground environments {CTU-CRAS-NORLAB} in the {DARPA} subterranean
  challenge,'' \emph{CoRR}, 2021. [Online]. Available:
  \url{https://arxiv.org/abs/2110.05911}
\BIBentrySTDinterwordspacing

\bibitem{Otsu2020IEEEAerospace}
K.~Otsu, S.~Tepsuporn, \emph{et~al.}, ``Supervised autonomy for
  communication-degraded subterranean exploration by a robot team,'' in
  \emph{IEEE Aerospace Conference}, 2020.

\bibitem{BaeRossiDavidoff2020VizAnalytics}
S.~S. Bae, F.~Rossi, \emph{et~al.}, ``A visual analytics approach to debugging
  cooperative, autonomous multi-robot systems’ worldviews,'' in \emph{IEEE
  Conference on Visual Analytics Science and Technology (VAST)}, 2020.

\bibitem{benton-etat-2012-OPTIC}
J.~Benton, A.~Coles, and A.~Coles, ``Temporal planning with preferences and
  time-dependent continuous costs,'' in \emph{Proceedings of the Twenty-Second
  International Conference on International Conference on Automated Planning
  and Scheduling}.\hskip 1em plus 0.5em minus 0.4em\relax AAAI Press, 2012.

\bibitem{rtsCombatStrategy}
A.~Khan, K.~Yang, \emph{et~al.}, ``A competitive combat strategy and tactics in
  rts games ai and starcraft,'' in \emph{Advances in Multimedia Information
  Processing -- PCM 2017}, B.~Zeng, Q.~Huang, \emph{et~al.}, Eds.\hskip 1em
  plus 0.5em minus 0.4em\relax Cham: Springer International Publishing, 2018.

\bibitem{plgrim}
\BIBentryALTinterwordspacing
S.~Kim, A.~Bouman, \emph{et~al.}, ``{PLGRIM:} hierarchical value learning for
  large-scale exploration in unknown environments,'' \emph{CoRR}, 2021.
  [Online]. Available: \url{https://arxiv.org/abs/2102.05633}
\BIBentrySTDinterwordspacing

\bibitem{terry2020object}
E.~Terry, X.~Lei, \emph{et~al.}, ``Object and gas source detection with robotic
  platforms in perceptually-degraded environments,'' in \emph{RSS Workshop:
  Robots in the Wild: Challenges in Deploying Robust Autonomy for Robotic
  Exploration}, 2020.

\bibitem{andwhite2012user}
\BIBentryALTinterwordspacing
J.~Huang, R.~White, and G.~Buscher, ``User see, user point: Gaze and cursor
  alignment in web search,'' in \emph{Proceedings of the 2012 Conference on
  Human Factors in Computing Systems (CHI)}, 2012. [Online]. Available:
  \url{https://www.microsoft.com/en-us/research/publication/user-see-user-point-gaze-and-cursor-alignment-in-web-search/}
\BIBentrySTDinterwordspacing

\end{thebibliography}

\end{document}